\documentclass[journal]{IEEEtai}

\usepackage[colorlinks,urlcolor=blue,linkcolor=blue,citecolor=blue]{hyperref}

\usepackage{color,array}

\usepackage{graphicx}
\usepackage{makecell}
\usepackage{array}
\usepackage{balance}
\usepackage{amsmath}
\usepackage{graphicx}
\graphicspath{{Figures/}}
\usepackage{pdfpages}
\usepackage{cite}
\usepackage{lipsum}
\usepackage{subfig} %should come after caption package if given, can't be used with caption2, defines subfloat
\usepackage[T1]{fontenc} % For less than and greater than sign

\usepackage{url}
\usepackage{float}

\usepackage{wrapfig}
\usepackage{lipsum}

\usepackage{multirow}
\usepackage{multicol}
\usepackage{longtable}
\usepackage{tabularx}
\usepackage{booktabs}
\usepackage{mathtools}

\usepackage{gensymb} %For degree sysmbol

\usepackage[font=scriptsize,labelfont=bf,tableposition=top]{caption}
\usepackage{amsmath}
\usepackage{amssymb}  % For Blackboard fonts, used in Set of natural numbers
\usepackage{amsfonts}
\usepackage{algorithm2e}
\SetAlCapFnt{\footnotesize}
\RestyleAlgo{ruled}
\usepackage{algpseudocode}

\newcolumntype{?}{!{\vrule width 1pt}}
\usepackage{hyperref}
\usepackage{url}
\usepackage{adjustbox}

\begin{document}

\title{Meteorology-Driven GPT4AP: A Multi-Task Forecasting LLM for Atmospheric Air Pollution in Data-Scarce Settings} 

\author{
\thanks{Manuscript received 26-Jan-2026; revised Month-Day-2026; accepted Month Day 2026.}  
}

\author{ Prasanjit~Dey,
     Soumyabrata~Dev,~\IEEEmembership{Member, IEEE}, and
      Bianca Schoen Phelan, 
 \thanks{Manuscript received 23-Jun-2024; revised Month-Day-2024; accepted Month Day 2024.}       
 \thanks{P. Dey is with ADAPT SFI Research Centre, School of Computer Science, Technological University Dublin, Ireland (e-mail: d22124678@mytudublin.ie).}
\thanks{S. Dev is with ADAPT SFI Research Centre, School of Computer Science, University College Dublin (e-mail: soumyabrata.dev@ucd.ie). }
\thanks{B.S. Phelan is with School of Computer Science, Technological University Dublin, Ireland (e-mail: bianca.schoenphelan@tudublin.ie).}% <-this % stops a space
}

\markboth{IEEE Transactions on Artificial Intelligence ,~Vol.~XX, No.~XX, XX~2026}%
{IEEE Transactions on Artificial Intelligence ,~Vol.~XX, No.~XX, XX~2026}

\maketitle

\begin{abstract}
Accurate forecasting of air pollution is important for environmental monitoring and policy support, yet data-driven models often suffer from limited generalization in regions with sparse observations. This paper presents Meteorology-Driven GPT for Air Pollution (GPT4AP), a parameter-efficient multi-task forecasting framework based on a pre-trained GPT-2 backbone and Gaussian rank-stabilized low-rank adaptation (rsLoRA). The model freezes the self-attention and feed-forward layers and adapts lightweight positional and output modules, substantially reducing the number of trainable parameters. GPT4AP is evaluated on six real-world air quality monitoring datasets under few-shot, zero-shot, and long-term forecasting settings. In the few-shot regime using 10\% of the training data, GPT4AP achieves an average MSE/MAE of 0.686/0.442, outperforming DLinear (0.728/0.530) and ETSformer (0.734/0.505). In zero-shot cross-station transfer, the proposed model attains an average MSE/MAE of 0.529/0.403, demonstrating improved generalization compared with existing baselines. In long-term forecasting with full training data, GPT4AP remains competitive, achieving an average MAE of 0.429, while specialized time-series models show slightly lower errors. These results indicate that GPT4AP provides a data-efficient forecasting approach that performs robustly under limited supervision and domain shift, while maintaining competitive accuracy in data-rich settings.

\end{abstract}

\begin{IEEEImpStatement}

This research advances the application of foundation models in environmental forecasting by introducing GPT4AP, a meteorology-driven large language model for multi-task air pollution prediction under data-scarce and cross-domain conditions. By enabling accurate few-shot and zero-shot forecasting with minimal labeled data and significantly reduced trainable parameters, GPT4AP addresses key limitations of existing data-intensive models. The framework supports reliable air quality prediction in regions with limited monitoring infrastructure, facilitating timely environmental assessment, policy planning, and public health protection. This work demonstrates the potential of parameter-efficient large models to enhance the scalability, accessibility, and robustness of AI-driven environmental monitoring systems.

\end{IEEEImpStatement}


\section{Introduction} \label{sec:introduction}

\IEEEPARstart{I}{n} recent years, the escalating severity of climate change has led to an increase in the frequency and intensity of extreme weather events, such as floods, heatwaves, and droughts. These events are closely linked to shifting concentrations of air pollutants, which further exacerbate their impact. According to the World Health Organization (WHO), nearly 99\% of the global population breathes air that exceeds WHO’s recommended pollution levels~\cite{kan2023air}. The intricate relationship between climate change and air pollution is well documented, as changes in atmospheric conditions directly influence the distribution and concentration of pollutants, including PM$_{2.5}$, PM$_{10}$, NO$_2$, SO$_2$, CO, CO$_2$, and O$_3$. As these pollutants interact with the climate system, they intensify the severity of weather extremes, creating a reinforcing cycle of environmental degradation~\cite{coughlan2023potential, rodell2023changing, nyayapathi2025comprehensive, hadei2025global, ofremu2025exploring, mu2026temperature}. This cycle has profound implications not only for public health, but also for long-term climate stability and ecosystem sustainability.

The rapid development of deep learning has significantly advanced air pollution modeling by enabling accurate predictions with reduced computational costs compared to traditional physics-based approaches~\cite{bi2023accurate, lam2023learning, camps2025artificial}. Unlike numerical models that explicitly simulate physical processes, data-driven models learn complex spatio-temporal relationships directly from observations, making them effective for pollution forecasting under dynamic environmental conditions. However, most existing deep learning approaches remain highly data-intensive~\cite{koldunov2024emerging, lilhore2025advanced, patel2025systematic} and rely on large-scale, well-curated datasets. For example, models such as FourCastNet~\cite{pathak2022fourcastnet} and AIFS~\cite{lang2024aifs} achieve impressive performance but require massive training corpora and are primarily designed for data-rich regimes. Although machine learning is increasingly explored for climate-related forecasting tasks~\cite{eyring2024pushing}, these models are often specialized for specific applications such as parameterization, bias correction, or climate impact assessment~\cite{henn2024machine, gupta2024machine, gregory2024machine, reder2025estimating}, and typically lack the flexibility to generalize across forecasting horizons, spatial domains, and data availability regimes.

Recently, foundation models have emerged as a powerful paradigm in artificial intelligence, enabling strong generalization across tasks and domains through large-scale pretraining. Pre-trained Large Language Models (LLMs), in particular, have demonstrated remarkable success in natural language processing and are increasingly being adapted for time-series analysis~\cite{zhou2023one}. However, their application in atmospheric and environmental sciences remains limited. Models such as AtmoRep~\cite{lessig2023atmorep} and Aurora~\cite{bodnar2024aurora} represent early efforts to introduce foundation models into weather and climate modeling, primarily focusing on nowcasting, downscaling, and bias correction. Despite their promise, these approaches often require substantial training data and are not explicitly designed to operate under data scarcity or across heterogeneous forecasting regimes. Moreover, the black-box nature of large neural models continues to raise concerns regarding interpretability, robustness, and generalization in environmental applications.

To address these challenges, we propose Meteorology-Driven GPT for Air Pollution (GPT4AP), a parameter-efficient, multi-task large language model designed specifically for air pollution forecasting in data-scarce settings. GPT4AP supports few-shot, zero-shot, and long-term forecasting within a unified framework, enabling consistent performance across varying data availability conditions. The model is built upon a pre-trained GPT-2 backbone, leveraging its rich representational capacity while freezing its self-attention and feed-forward layers to prevent overfitting and reduce computational cost. Adaptation is achieved through lightweight, trainable modules based on Gaussian rank-stabilized low-rank adaptation (rsLoRA), applied to custom positional embeddings and the prediction head. This design dramatically reduces the number of trainable parameters while preserving strong generalization across stations, horizons, and forecasting regimes. A rank-dependent Gaussian scaling mechanism further stabilizes optimization and inference across different temporal horizons.

We evaluate GPT4AP on six real-world air quality monitoring datasets under three forecasting settings: few-shot learning with limited training data, zero-shot cross-station transfer without target-domain fine-tuning, and long-term forecasting using full historical data. The experimental results demonstrate that GPT4AP consistently outperforms state-of-the-art time-series forecasting models in few-shot and zero-shot settings, highlighting its strong sample efficiency and cross-domain generalization capability. In long-term forecasting, GPT4AP achieves competitive performance comparable to specialized time-series models, while maintaining greater robustness across diverse monitoring locations and horizons.

The main contributions of this work are summarized as follows:

\begin{itemize}
\item We introduce GPT4AP, the first meteorology-driven, pre-trained LLMs specifically designed for multi-task air pollution forecasting under data-scarce conditions, supporting few-shot, zero-shot, and long-term prediction within a unified architecture.
\item We propose Gaussian rank-stabilized low-rank adaptation (rsLoRA) as a parameter-efficient transfer learning mechanism for environmental time-series forecasting, enabling substantial reductions in trainable parameters without sacrificing predictive performance.
\item Through extensive experiments on six real-world monitoring stations, we demonstrate that GPT4AP achieves state-of-the-art performance in few-shot and zero-shot forecasting, and competitive performance in long-term forecasting, establishing it as a robust and scalable foundation for data-efficient environmental modeling.
\end{itemize}

\section{Related Work}

Air pollution and climate forecasting have traditionally relied on numerical simulations and physical modeling to represent atmospheric transport, chemical reactions, and emission processes. These approaches remain essential for understanding the underlying mechanisms of pollution dynamics, but they are computationally expensive and often impractical for real-time or large-scale deployment. As a result, data-driven methods have become increasingly important for operational air quality forecasting, smart city applications, and environmental monitoring.

\textbf{Physics-Based Approaches:} Classical air pollution models are derived from physical and chemical laws governing atmospheric processes, including advection, diffusion, deposition, and chemical transformation~\cite{arystanbekova2004application, liu2005prediction, mohammadi2025air, skiba2025mathematical}. While these models offer strong physical interpretability, they typically require high computational resources, accurate emission inventories, and detailed boundary conditions, limiting their scalability and applicability in many regions. Consequently, their use in high-resolution, real-time, or data-scarce scenarios remains challenging~\cite{bracco2025machine}.

\textbf{Data-Driven and Hybrid Approaches:} Data-driven models leverage historical pollution and meteorological data to learn spatio-temporal relationships directly from observations~\cite{su2023novel, wang2022air, liang2023airformer, mulomba2025applying, cui2026air}. Deep learning architectures, including recurrent networks, attention-based models, and transformers, have shown strong performance in capturing complex nonlinear dependencies in air pollution time series. More recently, hybrid approaches have been proposed that integrate physical constraints into learning-based models to improve robustness and interpretability~\cite{wang2022traffic, ji2022stden, mohammadshirazi2023novel, rad2025predictive, khatibi2025advancing, saravana2026transformers}. These frameworks typically incorporate physical regularization, state-space formulations, or energy-based constraints into neural models.

Despite their success, most existing data-driven and hybrid methods remain heavily dependent on large volumes of labeled data and are often tailored to specific forecasting horizons, locations, or tasks. This limits their applicability in regions with sparse monitoring infrastructure and reduces their ability to generalize across stations, climates, and forecasting regimes. In contrast, foundation models and large pre-trained architectures offer a promising direction for improving generalization and transferability. Although LLMs and foundation models have recently been explored in weather and climate science~\cite{lessig2023atmorep, bodnar2024aurora}, their use in air pollution forecasting, particularly under data-scarce conditions remains largely unexplored.

Our work differs from existing approaches in two key aspects. First, we introduce a meteorology-driven, pre-trained LLM architecture specifically adapted for air pollution forecasting across few-shot, zero-shot, and long-term regimes within a unified framework. Second, we propose a parameter-efficient adaptation strategy based on Gaussian rsLoRA, which enables effective transfer learning with a minimal number of trainable parameters. This design allows GPT4AP to operate reliably in data-scarce environments while maintaining competitive performance in data-rich settings, addressing a critical gap in current air quality forecasting research.

\section{Methodology}

\subsection{Air Pollution Prediction Problem}

The goal of air pollution prediction is to forecast future PM$_{2.5}$ concentrations using historical pollutant concentrations (PM$_{2.5}$, PM$_{10}$, SO$_2$, NO$_2$, CO, and O$_3$) and meteorological observations (temperature, air pressure, dew point, and rainfall) collected from $K$ monitoring sites. We consider the set of monitoring sites $\{v_1,\dots,v_K\}$ and treat each site as providing an independent multivariate time series of observations. At each time step $t$, let $\mathbf{x}_t\in\mathbb{R}^{K\times 6}$ denote the pollutant observation matrix and $\mathbf{m}_t\in\mathbb{R}^{K\times 4}$ denote the meteorological observation matrix, where the $i$-th row of $\mathbf{x}_t$ and $\mathbf{m}_t$ corresponds to the measurements at site $v_i$. In all experiments, we use a fixed lookback window of length $T=36$ time steps (hours), which is kept constant across all monitoring sites, forecasting horizons, and data availability regimes. The historical observations are stacked as $\mathbf{x}_{t-T+1:t}\in\mathbb{R}^{T\times K\times 6}$ and $\mathbf{m}_{t-T+1:t}\in\mathbb{R}^{T\times K\times 4}$. Let $\mathbf{y}_t\in\mathbb{R}^{K\times 1}$ denote the ground-truth PM$_{2.5}$ concentration vector at time $t$, corresponding to the PM$_{2.5}$ channel of $\mathbf{x}_t$. The objective is to learn a forecasting function $h(\cdot)$ that predicts future PM$_{2.5}$ concentrations for all sites over a horizon of $\delta$ time steps:
\begin{equation}
\hat{\mathbf{y}}_{t+1:t+\delta} = h\!\left(\mathbf{x}_{t-T+1:t},\, \mathbf{m}_{t-T+1:t}\right)
\end{equation}
where $\hat{\mathbf{y}}_{t+1:t+\delta}\in\mathbb{R}^{\delta\times K\times 1}$ denotes the predicted PM$_{2.5}$ values for the next $\delta$ time steps.

% Although the formulation is written over a collection of $K$ sites, in this work we do not explicitly model inter-site relationships and do not use any graph-based operations. Instead, each model instance is trained on data from a single station, and the station index is therefore used only to denote different datasets and is omitted in the subsequent architecture description for notational simplicity.

\begin{figure*}[t]
    \centering
    \includegraphics[width=0.85\linewidth]{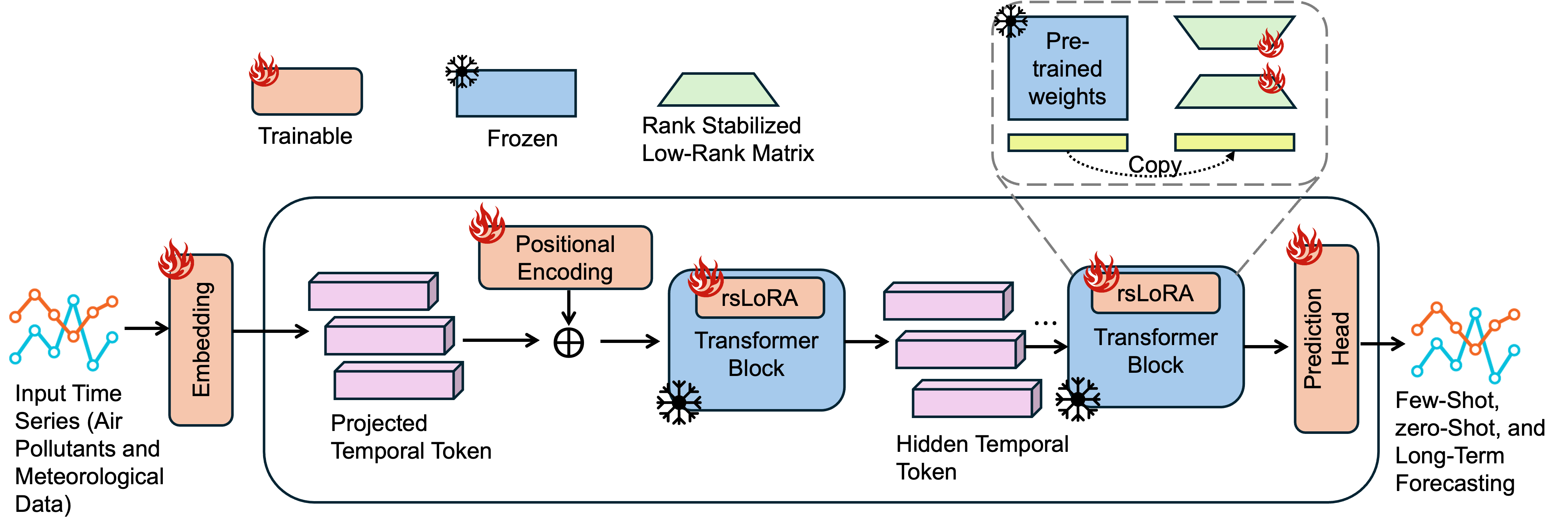}
    \caption{Overview of GPT4AP. Multivariate air-pollution and meteorological time series are embedded and projected into temporal tokens. Trainable positional encoding is added before feeding tokens into a frozen GPT-2 backbone. Parameter-efficient Gaussian rsLoRA is applied to lightweight adaptation modules, while the pretrained transformer weights remain frozen. A prediction head produces outputs for few-shot, long-term, and zero-shot forecasting across multiple horizons.}
    \label{fig:gpt4ap_framework}
\end{figure*}

\subsection{Overview of GPT4AP}
As illustrated in Figure~\ref{fig:gpt4ap_framework}, GPT4AP is a parameter-efficient, multi-task forecasting framework built upon a pre-trained GPT-2 backbone. Given historical air pollutant and meteorological observations, GPT4AP predicts future PM$_{2.5}$ concentrations under three forecasting regimes: few-shot, long-term, and zero-shot forecasting. The core design of GPT4AP is to freeze the computationally expensive self-attention and feed-forward layers of the pretrained LLM and adapt the model using lightweight trainable components based on Gaussian rsLoRA. This design significantly reduces the number of trainable parameters while preserving strong generalization across datasets, horizons, and tasks.

\subsection{Input Representation and Temporal Tokenization}
At each time step $t$, we form the multivariate input by concatenating pollutant and meteorological variables for each monitoring station:
\begin{equation}
\mathbf{s}_t = \left[\mathbf{x}_t \, \| \, \mathbf{m}_t\right] \in \mathbb{R}^{d}, \quad d = 10
\end{equation}
where $\mathbf{x}_t$ contains pollutant variables (PM$_{2.5}$, PM$_{10}$, SO$_2$, NO$_2$, CO, and O$_3$) and $\mathbf{m}_t$ contains meteorological variables (temperature, air pressure, dew point, and rainfall). In all experiments, we use a fixed lookback window of length $T = 36$ time steps (hours). The resulting input sequence is given by
\begin{equation}
\mathbf{S}_{t-T+1:t} = \left[\mathbf{s}_{t-T+1}, \dots, \mathbf{s}_t\right] \in \mathbb{R}^{T \times d}
\end{equation}

\textbf{Normalization.} To stabilize training across heterogeneous stations and seasonal variability, we apply Z-score normalization independently to each variable using statistics computed within the same input window:
\begin{equation}
\tilde{\mathbf{S}}_{t-T+1:t} =
\frac{\mathbf{S}_{t-T+1:t} - \boldsymbol{\mu}_{t}}{\boldsymbol{\sigma}_{t}}
\end{equation}
where $\boldsymbol{\mu}_{t}, \boldsymbol{\sigma}_{t} \in \mathbb{R}^{d}$ denote the mean and standard deviation computed over the temporal dimension of the lookback window. This normalization is applied consistently during training, validation, and testing. De-normalization is applied to the final PM$_{2.5}$ predictions to recover values in the original scale. %The effect of this per-window normalization is further analyzed via a station-level normalization variant based on training-split statistics in the ablation study.

\textbf{Patching and Temporal Tokens.}
Following normalization, the input sequence $\tilde{\mathbf{S}}_{t-T+1:t} \in \mathbb{R}^{T \times d}$ is segmented using a sliding-window patching scheme with patch length $P$ and stride $s$. Specifically, overlapping temporal patches are extracted as
\begin{equation}
\tilde{\mathbf{S}}^{(i)} = \tilde{\mathbf{S}}_{(i-1)s+1:(i-1)s+P}, \quad i = 1, \dots, N
\end{equation}
where the number of patches is given by
\begin{equation}
N = \left\lfloor \frac{T - P}{s} \right\rfloor + 1
\end{equation}
Each patch is flattened and linearly projected into the GPT hidden dimension $D$:
\begin{equation}
\mathbf{z}_i = \mathbf{W}_{\mathrm{proj}}
\left(\mathrm{Flatten}\left(\tilde{\mathbf{S}}^{(i)}\right)\right)
\end{equation}
forming a sequence of temporal tokens
$\mathbf{Z} = [\mathbf{z}_1, \dots, \mathbf{z}_N] \in \mathbb{R}^{N \times D}$,
which is provided as input to the transformer backbone.

\subsection{Trainable Positional Encoding}
Unlike text, time series require explicit temporal ordering. We use a learnable positional embedding $\mathbf{E}_{\text{pos}} \in \mathbb{R}^{N \times D}$ is added to the token sequence:
\begin{equation}
\mathbf{H}_0 = \mathbf{Z} + \mathbf{E}_{\text{pos}}
\end{equation}
To keep adaptation parameter-efficient, GPT4AP applies rsLoRA to the positional module, so that most pretrained weights remain unchanged while positional adaptation remains stable across ranks.

\subsection{Frozen GPT-2 Backbone}
Let $\mathcal{F}_{\text{GPT2}}(\cdot)$ denote the stacked transformer blocks of GPT-2. GPT4AP keeps all self-attention and feed-forward parameters frozen during training:
\begin{equation}
\mathbf{H}_L = \mathcal{F}_{\text{GPT2}}(\mathbf{H}_0), \quad
\frac{\partial \mathcal{F}_{\text{GPT2}}}{\partial \theta_{\text{GPT2}}}=0
\end{equation}
where $L$ is the number of transformer layers and $\theta_{\text{GPT2}}$ denotes pretrained parameters. This transfer-learning strategy enables GPT4AP to leverage rich temporal representations learned during pretraining while avoiding overfitting in data-scarce regimes.

\subsection{Gaussian Rank-Stabilized LoRA (rsLoRA)}
To adapt GPT4AP without updating the backbone, we use rsLoRA, a low-rank decomposition with a Gaussian initialization and rank-dependent scaling. For a target weight matrix $W_0 \in \mathbb{R}^{a \times b}$, rsLoRA applies:
\begin{equation}
W = W_0 + \Delta W, \qquad \Delta W = \beta_r YX
\end{equation}
where $Y \in \mathbb{R}^{a \times r}$ and $X \in \mathbb{R}^{r \times b}$ are trainable matrices and $r \ll \min(a,b)$ is the rank. We initialize $X$ with i.i.d.\ Gaussian entries and set $Y=0$ so that $\Delta W=0$ at the start of training. The scaling factor is:
\begin{equation}
\beta_r = \frac{\alpha}{\sqrt{r}}
\end{equation}
which stabilizes training dynamics across ranks by preventing activation and gradient magnitudes from exploding or vanishing as $r$ increases, while the rank itself still controls the expressive capacity of the adaptation.

\subsection{Prediction Head for Multi-Horizon Forecasting}

The final hidden token representations are summarized using mean pooling over the temporal tokens:
\begin{equation}
\mathbf{u} = \frac{1}{N}\sum_{i=1}^{N} \mathbf{H}_L^{(i)}
\end{equation}

A single lightweight prediction head maps $\mathbf{u}$ to a $\delta$-step forecast in one forward pass:
\begin{equation}
\hat{\mathbf{y}}_{t+1:t+\delta} = f_{\text{out}}(\mathbf{u})
\end{equation}
where $f_{\text{out}}$ is implemented as a multi-layer perceptron with output dimension $\delta$ and is augmented with rsLoRA for parameter efficiency. The same pooling and prediction head are used for all forecasting horizons.

\subsection{Training Protocols for Long-Term, Few-Shot, and Zero-Shot Forecasting}
GPT4AP supports multiple forecasting regimes without changing the architecture: in the long-term forecasting setting, we train a separate instance of GPT4AP on the full available training split for each station and evaluate on the test split for horizons $\delta\in\{24,36,48,60\}$. In the few-shot forecasting setting, we fine-tune only the rsLoRA parameters and lightweight modules using a small fraction of data (e.g., 10\%), while keeping the GPT-2 backbone frozen in order to emphasize sample efficiency. In the zero-shot forecasting setting, we train GPT4AP on a single source station and directly evaluate the same model on an unseen target station without any additional fine-tuning to test cross-station generalization. For all settings, the objective function is:
\begin{equation}
\mathcal{L} = \frac{1}{\delta}\sum_{k=1}^{\delta}\|y_{t+k}-\hat{y}_{t+k}\|_2^2
\end{equation}

The overall training and inference procedures of GPT4AP are summarized in Algorithm~\ref{alg:gpt4ap}.

%==================== Algorithm: GPT4AP Training & Inference (Revised) ====================
\begin{algorithm}[t]
\footnotesize
\caption{\footnotesize GPT4AP: Parameter-Efficient Training and Inference with rsLoRA.}
\label{alg:gpt4ap}
\DontPrintSemicolon
\SetKwInOut{Input}{Input}\SetKwInOut{Output}{Output}

\Input{Multivariate sequence $\mathbf{S}_{t-T+1:t}\in\mathbb{R}^{T\times d}$ (pollutants+meteorology), horizon $\delta$, patch length $P$, stride $s$, pretrained GPT-2 backbone $\mathcal{F}_{\text{GPT2}}$, rsLoRA rank $r$, scaling $\beta_r=\alpha/\sqrt{r}$, learning rate $\eta$}
\Output{Forecast $\hat{\mathbf{y}}_{t+1:t+\delta}$ for PM$_{2.5}$; trained rsLoRA parameters $\{X,Y\}$ and lightweight modules}

\vspace{2pt}
\textbf{Initialize:} Freeze GPT-2 parameters $\theta_{\text{GPT2}}$;\;
Initialize trainable projection $\mathbf{W}_{\text{proj}}$, positional embedding $\mathbf{E}_{\text{pos}}$, and output head $f_{\text{out}}$;\;
Initialize rsLoRA parameters for each adapted module (positional module and output head):
$X\sim\mathcal{N}(0,\sigma^2)$ and set $Y\leftarrow \mathbf{0}$ (so $\Delta W=\beta_rYX=\mathbf{0}$ at start).\;

\vspace{2pt}
\BlankLine
\textbf{Training Phase (Long-term / Few-shot)}\;
\For{each minibatch $\{\mathbf{S}_{t-T+1:t}, \mathbf{y}_{t+1:t+\delta}\}$}{
    \textbf{Normalize (per-window):} $\tilde{\mathbf{S}} \leftarrow (\mathbf{S}-\boldsymbol{\mu}_{t})/\boldsymbol{\sigma}_{t}$ \tcp*{$\boldsymbol{\mu}_{t},\boldsymbol{\sigma}_{t}$ computed over the input window per variable}
    \textbf{Sliding-window patching:} $N \leftarrow \left\lfloor\frac{T-P}{s}\right\rfloor+1$\;
    \For{$i=1$ \KwTo $N$}{
        $\tilde{\mathbf{S}}^{(i)} \leftarrow \tilde{\mathbf{S}}_{(i-1)s+1:(i-1)s+P}$\;
        $\mathbf{z}_i \leftarrow \mathbf{W}_{\text{proj}}\!\left(\mathrm{Flatten}\!\left(\tilde{\mathbf{S}}^{(i)}\right)\right)$\;
    }
    $\mathbf{Z}\leftarrow[\mathbf{z}_1,\dots,\mathbf{z}_N]$\;

    \textbf{rsLoRA (positional):} $\mathbf{E}_{\text{pos}} \leftarrow \mathbf{E}_{\text{pos}}^{0} + \beta_r\,Y_{\text{pos}}X_{\text{pos}}$\;
    \textbf{Position:} $\mathbf{H}_0 \leftarrow \mathbf{Z} + \mathbf{E}_{\text{pos}}$\;
    \textbf{Frozen Backbone:} $\mathbf{H}_L \leftarrow \mathcal{F}_{\text{GPT2}}(\mathbf{H}_0)$ \tcp*{$\nabla_{\theta_{\text{GPT2}}}=0$}
    \textbf{Pooling:} $\mathbf{u}\leftarrow \frac{1}{N}\sum_{i=1}^{N}\mathbf{H}_L^{(i)}$ \tcp*{mean pooling over tokens}\;

    \textbf{rsLoRA (output):} $W_{\text{out}} \leftarrow W_{\text{out}}^{0} + \beta_r\,Y_{\text{out}}X_{\text{out}}$\;
    \textbf{Predict (direct multi-horizon):} $\hat{\mathbf{y}}_{t+1:t+\delta} \leftarrow f_{\text{out}}(\mathbf{u}; W_{\text{out}})$\;
    \textbf{Loss:} $\mathcal{L}\leftarrow \frac{1}{\delta}\sum_{k=1}^{\delta}\|y_{t+k}-\hat{y}_{t+k}\|_2^2$\;
    \textbf{Update:} backpropagate and update only trainable parameters
    $\{\mathbf{W}_{\text{proj}}, \mathbf{E}_{\text{pos}}, f_{\text{out}}, X_{\text{pos}},Y_{\text{pos}},X_{\text{out}},Y_{\text{out}}\}$ using Adam with step size $\eta$\;
}

\vspace{2pt}
\BlankLine
\textbf{Inference Phase (Long-term / Few-shot / Zero-shot)}\;
\Input{$\mathbf{S}_{t-T+1:t}$}\;
\textbf{Normalize (per-window):} $\tilde{\mathbf{S}} \leftarrow (\mathbf{S}-\boldsymbol{\mu}_{t})/\boldsymbol{\sigma}_{t}$\;
\textbf{Sliding-window patching and tokenization:} compute $N=\left\lfloor\frac{T-P}{s}\right\rfloor+1$ and form $\mathbf{Z}\in\mathbb{R}^{N\times D}$ as in training\;
\textbf{rsLoRA (positional):} $\mathbf{E}_{\text{pos}} \leftarrow \mathbf{E}_{\text{pos}}^{0} + \beta_r\,Y_{\text{pos}}X_{\text{pos}}$\;
$\mathbf{H}_0 \leftarrow \mathbf{Z} + \mathbf{E}_{\text{pos}}$\;
$\mathbf{H}_L \leftarrow \mathcal{F}_{\text{GPT2}}(\mathbf{H}_0)$\;
\textbf{Pooling:} $\mathbf{u}\leftarrow \frac{1}{N}\sum_{i=1}^{N}\mathbf{H}_L^{(i)}$\;
\textbf{rsLoRA (output):} $W_{\text{out}} \leftarrow W_{\text{out}}^{0} + \beta_r\,Y_{\text{out}}X_{\text{out}}$\;
$\hat{\mathbf{y}}_{t+1:t+\delta} \leftarrow f_{\text{out}}(\mathbf{u}; W_{\text{out}})$\;
\Return $\hat{\mathbf{y}}_{t+1:t+\delta}$\;

\end{algorithm}
%===============================================================================
\section{Experimental Setup}
\subsection{Dataset Description}

We evaluate GPT4AP using air pollution and meteorological data collected from six monitoring stations in China~\cite{dey2023combinedeepnet}: Aoti Zhongxin (AZ), Dongsi (DS), Shunyicheng (SY), Tiantan (TT), Haidian Wanliu (HW), and Wanshou Xigong (WX). The dataset spans from March 1, 2013, to February 28, 2017, and contains hourly measurements of ten variables: six air pollutants (PM$_{2.5}$, PM$_{10}$, SO$_2$, NO$_2$, CO, and O$_3$) and four meteorological factors (temperature, air pressure, dew point, and rainfall). After preprocessing and removing incomplete or inconsistent records, we obtain approximately 26,300 hours of data per station for training and 8,760 hours for testing, corresponding to a standard three-year training and one-year testing split.

\subsection{Baseline Models and Implementation Details}

We compare GPT4AP against six state-of-the-art time-series forecasting models: DLinear~\cite{zeng2023transformers}, ETSformer~\cite{woo2022etsformer}, FiLM~\cite{zhou2022film}, Informer~\cite{zhou2021informer}, Pyraformer~\cite{liu2022pyraformer}, and the standard Transformer~\cite{wen2022transformers}. All baseline models are evaluated under identical experimental settings to ensure a fair comparison. Experiments are conducted under two data regimes: a few-shot setting using only 10\% of the available training data, and a long-term setting using the full training dataset. GPT4AP is implemented in PyTorch~1.13.1 and trained on an NVIDIA A100 GPU. Unless otherwise specified, GPT4AP employs rank-32 Gaussian rsLoRA adapters. We optimize all models using the Adam optimizer and report mean squared error (MSE) and mean absolute error (MAE). Table~\ref{tab:hyperparameters} summarizes the key hyperparameters used in our experiments.

\begin{table}[htbp]
\centering
\caption{Hyperparameter settings for GPT4AP.}
\label{tab:hyperparameters}
\begin{tabular}{ll}
\hline
\textbf{Hyperparameter} & \textbf{Value} \\
\hline
Backbone & GPT-2 (pre-trained) \\
Fine-tuning method & rsLoRA \\
rsLoRA initialization & Gaussian \\
Optimizer & Adam \\
Rank & 32\\
Learning rate & $1 \times 10^{-4}$ \\
Batch size & 16 \\
Training epochs & 10 \\
Early stopping patience & 3 \\
Loss function & MSE \\
Prediction horizon ($H$) & 24, 36, 48, 60 \\
Training data usage & 10\%, 100\% \\
Model dimension ($d_{\text{model}}$) & 768 \\
Number of heads ($n_{\text{heads}}$) & 4 \\
FFN dimension ($d_{\text{ff}}$) & 768 \\
GPT layers & 6 \\
Encoder layers & 3 \\
Dropout rate & 0.2 \\
Patch size (P) & 24 \\
Input sequence length (T) & 36 \\
Prediction horizons & 24, 36, 48, 60 \\

\hline
\end{tabular}
\end{table}

\section{Experimental Results}

\subsection{Few-shot Forecasting}

\begin{table}[!ht]
\centering
\caption{Comparison of few-shot forecasting performance across six monitoring stations. All models are trained with 10\% of the training data and evaluated at horizons of 24, 36, 48, and 60 hours. Results are reported as MSE/MAE. \textbf{Bold} values indicate the best performance, and \underline{underlined} values indicate the second-best.}
\label{tab:few_sota}
\scriptsize
\setlength{\tabcolsep}{2pt}
\resizebox{\linewidth}{!}{
\begin{tabular}{l|c|c|c|c|c|c|c|c}
\toprule
Dataset & Horizon & GPT4AP & DLinear & ETSformer & FiLM & Informer & Pyraformer & Transformer \\
\midrule

\multirow{4}{*}{AZ}
 & 24 & 0.546/0.338 & 0.628/0.491 & 0.589/0.440 & 0.657/0.442 & 0.853/0.592 & 0.724/0.544 & 0.766/0.556\\
 & 36 & 0.633/0.431 & 0.692/0.525 & 0.685/0.492 & 0.701/0.466 & 1.147/0.742 & 0.824/0.585 & 0.874/0.615\\
 & 48 & 0.686/0.457 & 0.724/0.543 & 0.756/0.528 & 0.747/0.486 & 1.014/0.718 & 0.930/0.642 & 0.938/0.658\\
 & 60 & 0.731/0.477 & 0.753/0.557 & 0.788/0.552 & 0.784/0.500 & 1.254/0.787 & 0.901/0.634 & 1.029/0.693\\
 & Avg & \textbf{0.649}/\textbf{0.426} & \underline{0.699}/0.529 & 0.705/0.503 & 0.722/\underline{0.474} & 1.067/0.710 & 0.845/0.601 &	0.902/0.631 \\
\midrule

\multirow{4}{*}{DS}
 & 24 & 0.662/0.400 & 0.725/0.497 & 0.694/0.447 & 0.788/0.458 & 0.896/0.571 & 0.786/0.530 & 0.803/0.535\\
 & 36 & 0.755/0.447 & 0.789/0.530 & 0.782/0.495 & 0.818/0.481 & 1.072/0.678 & 0.857/0.585 & 0.927/0.608\\
 & 48 & 0.817/0.474 & 0.826/0.548 & 0.856/0.529 & 0.871/0.503 & 1.159/0.699 & 0.933/0.613 & 0.982/0.624\\
 & 60 & 0.874/0.496 & 0.856/0.562 & 0.903/0.555 & 0.911/0.519 & 1.201/0.720 & 0.967/0.619 & 1.095/0.693\\
 & Avg & \textbf{0.777}/\textbf{0.454} & \underline{0.799}/0.534 & 0.809/0.507& 0.847/\underline{0.490} & 1.082/0.667 & 0.886/0.587 & 0.952/0.615\\
\midrule

\multirow{4}{*}{SY}
 & 24 & 0.535/0.414 & 0.609/0.509 & 0.567/0.461 & 0.637/0.468 & 0.733/0.558 & 0.647/0.524 & 0.686/0.551\\
 & 36 & 0.622/0.457 & 0.668/0.540 & 0.639/0.503 & 0.690/0.486 & 0.882/0.655 & 0.731/0.594 & 0.764/0.600\\
 & 48 & 0.670/0.480 & 0.699/0.556 & 0.706/0.543 & 0.729/0.503 & 0.857/0.659 & 0.767/0.590 & 0.830/0.636\\
 & 60 & 0.709/0.495 & 0.725/0.569 & 0.747/0.560 & 0.764/0.517 & 0.961/0.679 & 0.782/0.611 & 0.865/0.647\\
 & Avg & \textbf{0.634}/\textbf{0.462} & 0.675/0.544 & \underline{0.665}/0.517 & 0.705/\underline{0.494} & 0.858/0.638 & 0.732/0.580 & 0.786/0.609\\
\midrule

\multirow{4}{*}{TT}
 & 24 & 0.659/0.403 & 0.731/0.505 & 0.711/0.463 & 0.781/0.459 & 0.927/0.574 & 0.851/0.562 & 0.886/0.577\\
 & 36 & 0.736/0.443 & 0.794/0.534 & 0.806/0.506 & 0.809/0.475 & 1.141/0.701 & 0.969/0.636 & 1.031/0.647\\
 & 48 & 0.794/0.468 & 0.827/0.546 & 0.872/0.543 & 0.856/0.493 & 1.099/0.704 & 1.040/0.647 & 1.071/0.670\\
 & 60 & 0.844/0.488 & 0.855/0.561 & 0.924/0.572 & 0.895/0.508 & 1.206/0.745 & 1.036/0.634 & 1.198/0.738\\
 & Avg & \textbf{0.758}/\textbf{0.451} & \underline{0.802}/0.537 & 0.828/0.521 & 0.835/\underline{0.484} & 1.093/0.681 & 0.974/0.620& 1.047/0.658\\
\midrule

\multirow{4}{*}{HW}
 & 24 & 0.491/0.380 & 0.568/0.479 & 0.519/0.428 & 0.591/0.430 & 0.668/0.523 & 0.624/0.503 & 0.418/0.351\\
 & 36 & 0.575/0.422 & 0.632/0.513 & 0.601/0.466 & 0.635/0.456 & 0.807/0.603 & 0.727/0.546 & 0.507/0.406\\
 & 48 & 0.626/0.447 & 0.666/0.531 & 0.687/0.514 & 0.678/0.475 & 0.871/0.619 & 0.819/0.606 & 0.562/0.431\\
 & 60 & 0.668/0.466 & 0.695/0.545 & 0.734/0.539 & 0.713/0.489 & 0.919/0.642 & 0.774/0.580 & 0.595/0.443\\
 & Avg & \underline{0.590}/\underline{0.429} & 0.640/0.517 & 0.635/0.487	& 0.654/0.463 & 0.816/0.597 & 0.736/0.559 & \textbf{0.521}/\textbf{0.408}\\
\midrule

\multirow{4}{*}{WX}
 & 24 & 0.609/0.385 & 0.682/0.483 & 0.652/0.443 & 0.730/0.438 & 0.900/0.579 & 0.891/0.579 & 0.849/0.568\\
 & 36 & 0.690/0.426 & 0.746/0.516 & 0.734/0.481 & 0.757/0.461 & 1.087/0.689 & 1.013/0.651 & 0.983/0.639\\
 & 48 & 0.747/0.452 & 0.781/0.530 & 0.813/0.525 & 0.803/0.481 & 1.118/0.723 & 1.010/0.644 & 1.071/0.692\\
 & 60 & 0.796/0.472 & 0.810/0.539 & 0.856/0.545 & 0.843/0.496 & 1.319/0.798 & 1.049/0.662 & 1.205/0.736\\
 & Avg & \textbf{0.711}/\textbf{0.434} & \underline{0.755}/0.517 & 0.764/0.499 & 0.783/\underline{0.469} & 1.106/0.697 & 0.991/0.634 & 1.027/0.659\\
 \midrule
 \multicolumn{2}{c|}{All Avg} & \textbf{0.686}/\textbf{0.442} & \underline{0.728}/0.530 & 0.734/0.505 & 0.758/\underline{0.479} & 1.004/0.665 & 0.861/0.597 &  0.872/0.596\\
\bottomrule
\end{tabular}
}
\end{table}

To evaluate model performance under data scarcity, we train all models using only 10\% of the available training data. Table~\ref{tab:few_sota} reports the detailed station-level performance across four forecasting horizons (24, 36, 48, and 60 hours). Overall, GPT4AP achieves the best performance on most stations and horizons, demonstrating strong few-shot learning capability and robust generalization. At Aotizhongxin, GPT4AP achieves the lowest average MSE and MAE (0.649/0.426), improving over the strongest baseline, DLinear (0.699/0.529), by 7.2\% in MSE and 19.5\% in MAE. At Dongsi, GPT4AP attains an average MSE of 0.777, outperforming DLinear (0.799) by 2.8\% and ETSformer (0.809) by 4.0\%. For Shunyi, GPT4AP records the lowest average MSE of 0.634, yielding a 4.7\% improvement over ETSformer (0.665) and a 6.1\% improvement over DLinear (0.675). At Tiantan, GPT4AP achieves an average MSE of 0.758, surpassing DLinear (0.802) by 5.5\% and FiLM (0.835) by 9.2\%. At Wanshou, GPT4AP obtains an average MSE of 0.711, improving over DLinear (0.755) by 5.8\% and over ETSformer (0.764) by 7.0\%. Wanliu presents a more nuanced case: while Transformer achieves the lowest average MSE (0.521), GPT4AP demonstrates competitive performance (0.590) and exhibits greater stability across longer horizons (36h-60h), where it consistently outperforms Transformer. 

Averaged across all stations and horizons, GPT4AP achieves the lowest overall MSE and MAE (0.686/0.442), outperforming DLinear (0.728/0.530), ETSformer (0.734/0.505), and FiLM (0.758/0.479). These results confirm that the combination of a pre-trained LLM backbone with Gaussian rank-stabilized low-rank adaptation enables GPT4AP to generalize effectively under data scarcity, making it well-suited for deployment in environments with limited labeled observations.

\subsection{Long-term Forecasting}

\begin{table}[!ht]
\centering
\caption{Comparison of long-term forecasting performance across six monitoring stations. All models are trained using the full training dataset and evaluated at horizons of 24, 36, 48, and 60 hours. Results are reported as MSE/MAE. \textbf{Bold} values indicate the best performance, and \underline{underlined} values indicate the second-best.}
\label{tab:long_sota}
\scriptsize
\setlength{\tabcolsep}{2pt}
\resizebox{\linewidth}{!}{
\begin{tabular}{l|c|c|c|c|c|c|c|c}
\toprule
Dataset & Horizon & GPT4AP & DLinear & ETSformer & FiLM & Informer & Pyraformer & Transformer \\
\midrule

\multirow{4}{*}{AZ}
 & 24 & 0.530/0.376 & 0.532/0.409 & 0.525/0.397 & 0.557/0.393 & 0.470/0.357 & 0.467/0.360 & 0.456/0.366\\
 & 36 & 0.613/0.419 & 0.605/0.455 & 0.611/0.443 & 0.634/0.433 & 0.565/0.403 & 0.548/0.403 & 0.607/0.428\\
 & 48 & 0.671/0.447 & 0.649/0.482 & 0.656/0.486 & 0.685/0.457 & 0.615/0.434 & 0.594/0.441 & 0.607/0.433\\
 & 60 & 0.712/0.467 & 0.684/0.503 & 0.705/0.515 & 0.726/0.476 & 0.644/0.449 & 0.650/0.464 & 0.688/0.467\\
 & Avg & 0.632/0.427 & 0.618/0.462 & 0.624/0.460	& 0.651/0.440 & \underline{0.574}/\textbf{0.411} & \textbf{0.565}/\underline{0.417} & 0.590/0.424\\
\midrule

\multirow{4}{*}{DS}
 & 24 & 0.636/0.389 & 0.636/0.420 & 0.612/0.398 & 0.668/0.405 & 0.578/0.385 & 0.579/0.390 & 0.597/0.395\\
 & 36 & 0.726/0.432 & 0.713/0.465 & 0.702/0.450 & 0.750/0.447 & 0.640/0.410 & 0.648/0.417 & 0.658/0.434\\
 & 48 & 0.790/0.462 & 0.761/0.494 & 0.770/0.494 & 0.808/0.473 & 0.690/0.439 & 0.700/0.443 & 0.752/0.452\\
 & 60 & 0.839/0.484 & 0.798/0.515 & 0.807/0.509 & 0.855/0.493 & 0.707/0.454 & 0.740/0.467 & 0.731/0.463\\
 & Avg & 0.748/0.442 & 0.727/0.474 & 0.723/0.463	& 0.770/0.455 & \textbf{0.654}/\textbf{0.422} & \underline{0.667}/\underline{0.429} & 0.685/0.436\\
\midrule

\multirow{4}{*}{SY}
 & 24 & 0.508/0.392 & 0.504/0.425 & 0.497/0.414 & 0.530/0.408 & 0.461/0.392 & 0.478/0.405 & 0.448/0.393\\
 & 36 & 0.592/0.435 & 0.575/0.469 & 0.572/0.447 & 0.609/0.448 & 0.518/0.420 & 0.514/0.438 & 0.559/0.436\\
 & 48 & 0.645/0.461 & 0.617/0.493 & 0.625/0.491 & 0.658/0.471 & 0.594/0.447 & 0.571/0.454 & 0.616/0.462\\
 & 60 & 0.684/0.480 & 0.651/0.513 & 0.681/0.519 & 0.698/0.489 & 0.613/0.457 & 0.600/0.470 & 0.642/0.480\\
 & Avg & 0.607/\underline{0.442} & 0.587/0.475 & 0.594/0.468	& 0.624/0.454 & \underline{0.547}/\textbf{0.429} & \textbf{0.541}/0.442 & 0.566/0.443\\
\midrule

\multirow{4}{*}{TT}
 & 24 & 0.632/0.388 & 0.633/0.421 & 0.625/0.409 & 0.659/0.403 & 0.585/0.383 & 0.614/0.383 & 0.615/0.390\\
 & 36 & 0.715/0.427 & 0.709/0.466 & 0.726/0.462 & 0.739/0.442 & 0.708/0.421 & 0.682/0.418 & 0.726/0.432\\
 & 48 & 0.778/0.456 & 0.755/0.491 & 0.798/0.496 & 0.793/0.465 & 0.730/0.446 & 0.721/0.443 & 0.752/0.460\\
 & 60 & 0.821/0.475 & 0.791/0.511 & 0.825/0.529 & 0.838/0.483 & 0.753/0.464 & 0.813/0.472 & 0.809/0.480\\
 & Avg & 0.737/0.437 & 0.722/0.472 & 0.744/0.474	& 0.757/0.448 & \textbf{0.694}/\textbf{0.429} & \underline{0.708}/\underline{0.429} & 0.726/0.441\\
\midrule

\multirow{4}{*}{HW}
 & 24 & 0.473/0.362 & 0.470/0.395 & 0.470/0.381 & 0.494/0.382 & 0.425/0.357 & 0.434/0.365 & 0.637/0.499\\
 & 36 & 0.557/0.401 & 0.542/0.439 & 0.534/0.423 & 0.569/0.422 & 0.483/0.381 & 0.506/0.409 & 0.697/0.532\\
 & 48 & 0.612/0.422 & 0.587/0.465 & 0.610/0.470 & 0.620/0.445 & 0.519/0.407 & 0.522/0.436 & 0.839/0.608\\
 & 60 & 0.654/0.432 & 0.621/0.486 & 0.650/0.496 & 0.659/0.463 & 0.548/0.429 & 0.580/0.458 & 0.906/0.650\\
 & Avg &	0.574/\underline{0.404} & 0.555/0.446 & 0.566/0.443 & 0.586/0.428 & \textbf{0.494}/\textbf{0.394} & \underline{0.511}/0.417 & 0.770/0.572\\
\midrule

\multirow{4}{*}{WX}
 & 24 & 0.588/0.372 & 0.596/0.406 & 0.588/0.391 & 0.616/0.388 & 0.550/0.372 & 0.548/0.378 & 0.542/0.370 \\
 & 36 & 0.671/0.413 & 0.666/0.449 & 0.660/0.441 & 0.691/0.427 & 0.615/0.397 & 0.642/0.419 & 0.641/0.416 \\
 & 48 & 0.731/0.442 & 0.711/0.475 & 0.739/0.482 & 0.744/0.452 & 0.662/0.425 & 0.667/0.428 & 0.724/0.441\\
 & 60 & 0.779/0.464 & 0.747/0.497 & 0.778/0.512 & 0.788/0.471 & 0.682/0.447 & 0.698/0.441 & 0.735/0.450\\
 & Avg & 0.692/0.423 & 0.680/0.457 & 0.691/0.457	& 0.710/0.435 & \textbf{0.627}/\textbf{0.410} & \underline{0.639}/\underline{0.417} & 0.661/0.419\\
 \midrule
 \multicolumn{2}{c|}{All Avg} & 0.665/0.429 & 0.648/0.464 & 0.657/0.461 & 0.683/0.443 & \textbf{0.598}/\textbf{0.416} & \underline{0.605}/\underline{0.425} & 0.666/0.456\\
\bottomrule
\end{tabular}
}
\end{table}

We evaluate long-term forecasting performance using the full training dataset and four prediction horizons (24, 36, 48, and 60 hours). Table~\ref{tab:long_sota} reports the station-level MSE and MAE results. In contrast to the few-shot setting, specialized time-series models generally achieve lower errors under data-rich conditions, while GPT4AP remains competitive and stable across all stations and forecasting horizons in the long-term setting. At Aotizhongxin, Informer achieves the lowest average MAE error (0.411), followed closely by Pyraformer (0.417), whereas GPT4AP attains an average MAE of 0.427. At Dongsi, Informer again yields the best performance with an average MAE of 0.422, compared to 0.442 for GPT4AP. Similarly, at Shunyi, Informer leads with 0.429 MAE, while GPT4AP records second best 0.442. At Tiantan, Informer (0.429) and Pyraformer (0.429) outperform GPT4AP (0.437), though the differences remain moderate. At Wanliu, Informer achieves the lowest error (0.394), followed by second best GPT4AP (0.404) and significantly outperforms the standard Transformer (0.572). At Wanshou, Informer again provides the best average MAE (0.410), with GPT4AP achieving 0.423. Averaged across all six stations and horizons, Informer attains the lowest overall MAE (0.416), followed by Pyraformer (0.425). GPT4AP achieves an overall average of 0.429, which is comparable to Dlinear (0.464), Transformer (0.456) and FiLM (0.443).

Although GPT4AP does not dominate under data-rich long-term conditions, it demonstrates strong robustness and consistency across diverse monitoring stations and prediction horizons. Unlike specialized architectures that are tuned for specific temporal dynamics, GPT4AP provides a unified forecasting framework that performs reliably across heterogeneous environments. These results highlight that while specialized time-series models retain an advantage when abundant data are available, GPT4AP remains a competitive and versatile alternative with substantially fewer trainable parameters and stronger generalization capabilities.

\subsection{zero-shot forecasting}

\begin{table}[!ht]
\centering
\caption{Zero-shot forecasting performance across six cross-station transfer directions. Models are trained on a source station and evaluated on a different target station without any target-domain fine-tuning. Results are reported as MSE/MAE over four forecasting horizons (24, 36, 48, and 60 hours). \textbf{Bold} values indicate the best performance, and \underline{underlined} values indicate the second-best.}
\label{tab:zero_sota}
\scriptsize
\setlength{\tabcolsep}{2pt}
\resizebox{\linewidth}{!}{
\begin{tabular}{l|c|c|c|c|c|c|c|c}
\toprule
Dataset & Horizon & GPT4AP & DLinear & ETSformer & FiLM & Informer & Pyraformer & Transformer \\
\midrule

\multirow{4}{*}{\makecell{AZ \\ $\rightarrow$ \\ DS}}
 & 24  & 0.455/0.351 & 0.536/0.553 & 0.522/0.556 & 0.574/0.563 & 0.623/0.594 & 0.442/0.488 & 0.640/0.585 \\
 & 36  & 0.502/0.380 & 0.616/0.590 & 0.591/0.579 & 0.691/0.612 & 0.679/0.617 & 0.510/0.537 & 0.646/0.620 \\
 & 48  & 0.541/0.403 & 0.690/0.622 & 0.686/0.621 & 0.795/0.654 & 0.627/0.596 & 0.572/0.576 & 0.588/0.567 \\
 & 60  & 0.583/0.424 & 0.753/0.649 & 0.789/0.660 & 0.894/0.694 & 0.719/0.677 & 0.618/0.617 & 0.601/0.596 \\
 & Avg & \textbf{0.520}/\textbf{0.390} & 0.649/0.604 & 0.647/0.604 & 0.739/0.631 & 0.662/0.621 & \underline{0.536}/\underline{0.555} & 0.619/0.592 \\
\midrule

\multirow{4}{*}{\makecell{DS \\ $\rightarrow$ \\ AZ}}
 & 24  & 0.420/0.366 & 0.617/0.562 & 0.587/0.552 & 0.663/0.571 & 0.700/0.610 & 0.536/0.509 & 0.689/0.596 \\
 & 36  & 0.495/0.401 & 0.711/0.600 & 0.668/0.579 & 0.799/0.623 & 0.742/0.620 & 0.596/0.552 & 0.685/0.594 \\
 & 48  & 0.525/0.422 & 0.800/0.634 & 0.792/0.637 & 0.921/0.666 & 0.745/0.612 & 0.703/0.636 & 0.716/0.590  \\
 & 60  & 0.581/0.445 & 0.884/0.665 & 0.911/0.671 & 1.040/0.707 & 0.825/0.677 & 0.717/0.617 & 0.749/0.632  \\
 & Avg & \textbf{0.505}/\textbf{0.409} & 0.753/0.615 & 0.740/0.610 & 0.856/0.642 & 0.753/0.630 & \underline{0.638}/\underline{0.579} & 0.710/0.603\\
\midrule

\multirow{4}{*}{\makecell{AZ \\ $\rightarrow$ \\ TT}}
 & 24  & 0.484/0.364 & 0.514/0.540 & 0.486/0.532 & 0.551/0.544 & 0.618/0.589 & 0.422/0.473 & 0.591/0.560 \\
 & 36  & 0.527/0.392 & 0.590/0.576 & 0.553/0.555 & 0.659/0.595 & 0.673/0.614 & 0.488/0.522 & 0.615/0.601 \\
 & 48  & 0.562/0.414 & 0.658/0.606 & 0.638/0.596 & 0.752/0.634 & 0.606/0.586 & 0.542/0.560 & 0.571/0.559 \\
 & 60  & 0.600/0.433 & 0.713/0.631 & 0.727/0.633 & 0.840/0.670 & 0.686/0.662 & 0.585/0.599 & 0.576/0.583 \\
 & Avg & \underline{0.543}/\textbf{0.401} & 0.619/0.588 & 0.601/0.579	& 0.701/0.611 & 0.646/0.613 & \textbf{0.509}/\underline{0.539} & 0.588/0.576\\
\midrule

\multirow{4}{*}{\makecell{TT \\ $\rightarrow$ \\ AZ}}
 & 24  & 0.450/0.376 & 0.618/0.562 & 0.582/0.535 & 0.664/0.572 & 0.722/0.619 & 0.533/0.507 & 0.697/0.586 \\
 & 36  & 0.497/0.407 & 0.712/0.601 & 0.686/0.587 & 0.802/0.624 & 0.766/0.640 & 0.588/0.544 & 0.717/0.612 \\
 & 48  & 0.549/0.430 & 0.801/0.635 & 0.773/0.623 & 0.925/0.667 & 0.742/0.607 & 0.703/0.603 & 0.700/0.586 \\
 & 60  & 0.573/0.444 & 0.884/0.665 & 0.906/0.670 & 1.045/0.709 & 0.827/0.680 & 0.728/0.602 & 0.763/0.642 \\
 & Avg & \textbf{0.517}/\textbf{0.414} & 0.754/0.616 & 0.737/0.604 & 0.859/0.643 & 0.764/0.637 & \underline{0.638}/\underline{0.564} & 0.719/0.607 \\
\midrule

\multirow{4}{*}{\makecell{DS \\ $\rightarrow$ \\ TT}}
 & 24  & 0.489/0.368 & 0.516/0.541 & 0.475/0.520 & 0.558/0.548 & 0.619/0.593 & 0.426/0.477 & 0.583/0.564 \\
 & 36  & 0.546/0.398 & 0.592/0.577 & 0.550/0.554 & 0.665/0.598 & 0.662/0.606 & 0.493/0.527 & 0.574/0.573 \\
 & 48  & 0.572/0.417 & 0.656/0.606 & 0.659/0.609 & 0.756/0.635 & 0.605/0.590 & 0.593/0.615 & 0.587/0.563 \\
 & 60  & 0.617/0.438 & 0.716/0.632 & 0.743/0.642 & 0.842/0.671 & 0.690/0.662 & 0.585/0.594 & 0.636/0.610 \\
 & Avg & \underline{0.556}/\textbf{0.405} & 0.620/0.589 & 0.607/0.581	& 0.705/0.613 & 0.644/0.613 & \textbf{0.524}/\underline{0.553} &	0.595/0.578 \\
\midrule

\multirow{4}{*}{\makecell{TT \\ $\rightarrow$ \\ DS}}
 & 24  & 0.477/0.360 & 0.539/0.554 & 0.496/0.524 & 0.582/0.568 & 0.660/0.606 & 0.446/0.491 & 0.637/0.577 \\
 & 36  & 0.520/0.392 & 0.618/0.591 & 0.604/0.583 & 0.701/0.615 & 0.698/0.628 & 0.507/0.535 & 0.627/0.612 \\
 & 48  & 0.560/0.412 & 0.689/0.622 & 0.683/0.619 & 0.803/0.657 & 0.624/0.597 & 0.602/0.596 & 0.594/0.571 \\
 & 60  & 0.581/0.426 & 0.756/0.650 & 0.801/0.667 & 0.901/0.697 & 0.719/0.673 & 0.614/0.593 & 0.678/0.633 \\
 & Avg & \textbf{0.535}/\textbf{0.398} & 0.651/0.604 &	0.646/0.598 & 0.747/0.634 & 0.675/0.626	& \underline{0.542}/\underline{0.554} & 0.634/0.598 \\
 \midrule
 \multicolumn{2}{c|}{All Avg} & \textbf{0.529}/\textbf{0.403} & 0.674/0.603 & 0.663/0.596 & 0.768/0.629 & 0.691/0.623 & \underline{0.565}/\underline{0.557} & 0.644/0.592\\
\bottomrule
\end{tabular}
}
\end{table}

We assess cross-station generalization by training models on one station and directly evaluating on a different target station without any target-domain fine-tuning. Table~\ref{tab:zero_sota} reports the zero-shot MSE/MAE results across six transfer directions and four forecasting horizons. Overall, GPT4AP achieves the lowest average error in five out of six transfer directions, demonstrating strong cross-station generalization under distribution shift. In the Aotizhong $\rightarrow$ Dongsi transfer, GPT4AP attains the best average performance (0.520/0.390), improving over the second-best baseline, Pyraformer (0.536/0.555), by 3.0\% in MSE and also substantially outperforming FiLM (0.739/0.631). For Dongsi $\rightarrow$ Aotizhong, GPT4AP again achieves the lowest average MSE (0.505), outperforming the strongest baseline, Pyraformer (0.638), by 20.8\%. Similarly, for Tiantan $\rightarrow$ Aotizhong, GPT4AP achieves the best average MSE of 0.517, yielding a 19.0\% improvement over Pyraformer (0.638) and a 31.4\% improvement over DLinear (0.754). For Dongsi $\rightarrow$ Tiantan, GPT4AP attains the best average MSE (0.556), outperforming Pyraformer (0.524) in MAE but not in MSE; the lowest MSE in this transfer direction is achieved by Pyraformer (0.524). For Tiantan $\rightarrow$ Dongsi, GPT4AP yields the best average MSE (0.535), slightly improving over Pyraformer (0.542) by 1.3\%. The only transfer direction where GPT4AP does not achieve the best average performance is Aotizhong $\rightarrow$ Tiantan, where Pyraformer attains the lowest average MSE (0.509) and GPT4AP records 0.543, while still outperforming several other baselines such as DLinear (0.619) and FiLM (0.701).

Averaged across all transfer directions, horizons, and stations, GPT4AP achieves the best overall zero-shot performance with an average of 0.529/0.403, outperforming DLinear (0.674/0.603), ETSformer (0.663/0.596), FiLM (0.768/0.629), Informer (0.691/0.623), Pyraformer (0.565/0.557), and Transformer (0.644/0.592). These results indicate that GPT4AP effectively captures transferable meteorological-pollution relationships that generalize across monitoring sites, making it particularly valuable for deployment in regions with limited or missing labeled data.

\subsection{Ablation Study: Rank Sensitivity of rsLoRA}

\begin{figure*}[t]
\centering
\includegraphics[width=0.8\linewidth]{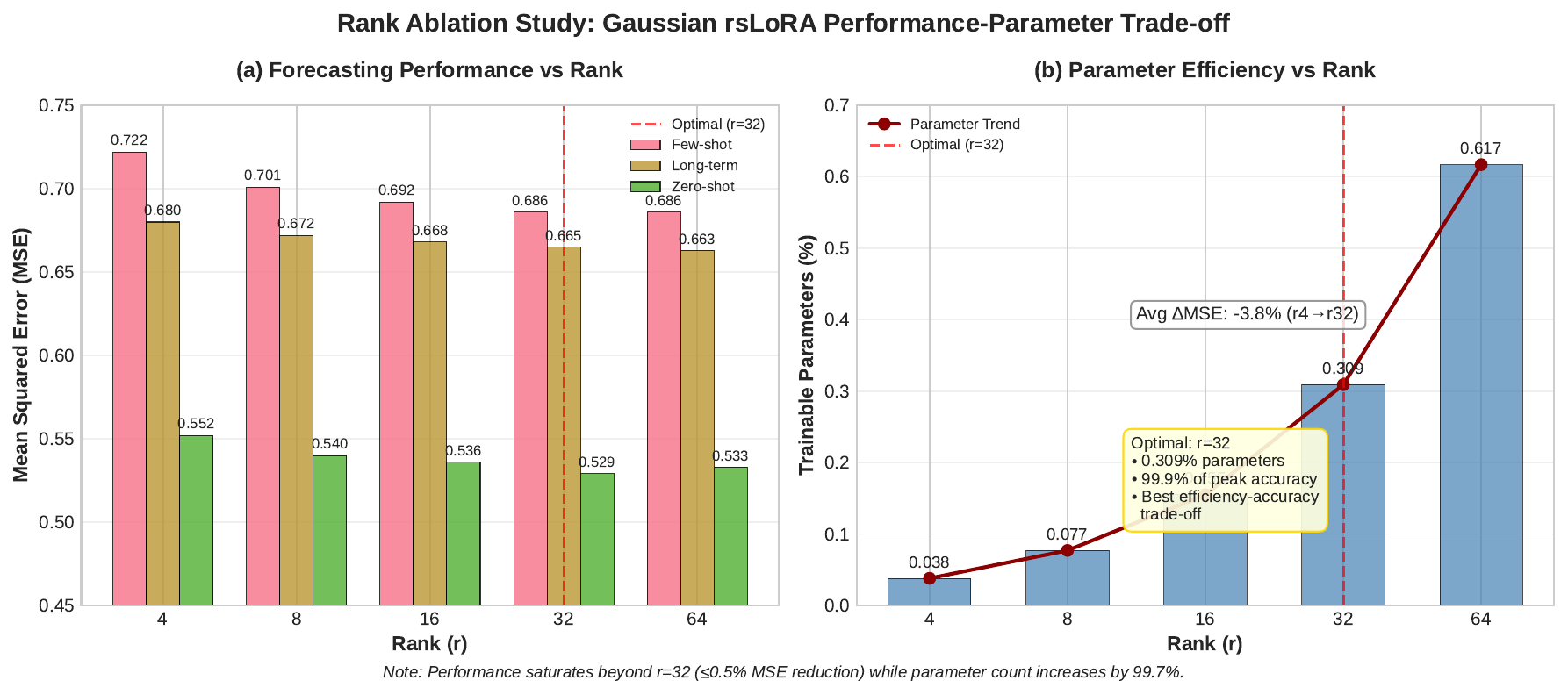}
\caption{Rank ablation study of Gaussian rsLoRA. (a) Forecasting error (MSE) decreases as rank increases and saturates beyond $r{=}32$. (b) Percentage of trainable parameters grows approximately linearly with rank. The best trade-off is achieved at $r{=}32$, which attains near-peak performance (99.9\% of the minimum MSE on average) while training only 0.309\% of the full model parameters.}
\label{fig:rank_ablation}
\end{figure*}

We conduct an ablation study to quantify the impact of the rsLoRA rank $r$ on forecasting accuracy and parameter efficiency. Figure~\ref{fig:rank_ablation} summarizes the average results over six datasets, three forecasting regimes (few-shot, long-term, and zero-shot), and all prediction horizons. We evaluate ranks $r \in \{4, 8, 16, 32, 64\}$. As shown in Figure~\ref{fig:rank_ablation}(a), increasing the rank consistently reduces MSE up to $r{=}32$ across all tasks. Moving from $r{=}4$ to $r{=}32$ yields average relative MSE reductions of 5.0\% (few-shot: 0.722 $\rightarrow$ 0.686), 2.2\% (long-term: 0.680 $\rightarrow$ 0.665), and 4.2\% (zero-shot: 0.552 $\rightarrow$ 0.529), corresponding to an average improvement of 3.8\% across tasks. In contrast, increasing the rank beyond $r{=}32$ provides negligible benefit: few-shot remains unchanged (0.686 $\rightarrow$ 0.686), long-term improves only marginally (0.665 $\rightarrow$ 0.663; $\approx$0.3\% relative), and zero-shot slightly degrades (0.529 $\rightarrow$ 0.533), indicating clear performance saturation beyond $r{=}32$. Figure~\ref{fig:rank_ablation}(b) shows that the trainable-parameter ratio increases nearly linearly with rank, from 0.038\% at $r{=}4$ to 0.309\% at $r{=}32$ and 0.617\% at $r{=}64$. Doubling the rank from $r{=}32$ to $r{=}64$ therefore increases the trainable parameters by 99.7\% while producing minimal (and task-dependent) accuracy changes. Overall, $r{=}32$ achieves 99.9\% of the minimum MSE on average while requiring only 0.309\% trainable parameters, making it the most efficient accuracy-cost trade-off for deployment.

Based on these findings, we set $r{=}32$ as the default configuration in all experiments. Detailed ablation results by station and horizon are reported in Appendix Tables I (few-shot), II (long-term), and III (zero-shot), which further confirm the saturation behavior beyond $r{=}32$.

\section{Discussion}

The experimental results highlight a clear distinction between data-scarce and data-rich forecasting regimes. GPT4AP consistently outperforms state-of-the-art baselines in few-shot and zero-shot settings, while remaining competitive but not dominant in long-term forecasting. This behavior reflects the strengths and limitations of leveraging a pre-trained large language model backbone for environmental time-series prediction. The strong performance of GPT4AP in few-shot and zero-shot forecasting can be attributed to its foundation-model prior and parameter-efficient adaptation. By freezing most of the pre-trained GPT-2 parameters and adapting only low-rank rsLoRA modules, GPT4AP benefits from rich, transferable representations learned from large-scale pretraining, allowing it to generalize effectively under limited supervision and cross-domain shifts. This enables the model to capture high-level meteorological-pollution relationships that are difficult to learn from small datasets alone, explaining its superior cross-station and low-data generalization. 

In contrast, in long-term forecasting with abundant data, specialized time-series architectures such as Informer and Pyraformer achieve lower errors. These models are explicitly designed to capture long-range temporal dependencies and periodic structures, which become learnable when sufficient data are available. GPT4AP, while flexible and robust, is not optimized for exploiting large volumes of homogeneous time-series data, which explains its slightly inferior performance in this regime. From a practical perspective, these results suggest that GPT4AP is particularly well suited for deployment in data-constrained environments, such as emerging regions, newly deployed sensor networks, or locations with incomplete monitoring infrastructure. In such scenarios, its ability to generalize across stations and operate with minimal labeled data provides a substantial advantage over conventional data-hungry models. The ablation study further demonstrates that GPT4AP achieves an efficient balance between accuracy and computational cost. A rank of $r=32$ enables near-peak performance with less than 0.31\% trainable parameters, making the model feasible for edge or resource-limited deployments while preserving predictive quality.

Nevertheless, this work has limitations. GPT4AP does not currently surpass specialized architectures in long-term forecasting under data-rich conditions, and the model has been evaluated on a limited set of stations within a single country. Future work will explore hybrid architectures that integrate explicit temporal modeling into the LLM framework, extend evaluation to broader geographic regions, and investigate multi-pollutant joint forecasting and uncertainty estimation. Overall, GPT4AP represents a promising step toward foundation-model-based environmental forecasting, demonstrating that large pre-trained models can be effectively adapted for atmospheric science, particularly in scenarios where data scarcity and generalization are the primary challenges.

\section{Conclusion}

This paper introduced GPT4AP, a meteorology-driven large language model for multi-task atmospheric air pollution forecasting under few-shot, zero-shot, and long-term settings. By leveraging a pre-trained GPT-2 backbone and a parameter-efficient Gaussian rsLoRA mechanism, GPT4AP achieves strong generalization while requiring only a small fraction of trainable parameters. Extensive experiments on six real-world air quality monitoring stations demonstrate that GPT4AP significantly outperforms state-of-the-art baselines in few-shot and zero-shot forecasting, where data availability is limited or where cross-station transfer is required. In long-term forecasting under data-rich conditions, GPT4AP remains competitive with specialized time-series models, although architectures explicitly designed for long-range temporal modeling retain an advantage. These results confirm that GPT4AP is particularly effective in data-scarce and distribution-shifted scenarios, making it well suited for deployment in practical environmental monitoring systems where labeled data are limited or unevenly distributed. The ablation study further shows that the proposed rsLoRA adaptation achieves an optimal balance between accuracy and efficiency at rank $r=32$, preserving near-peak forecasting performance while reducing the number of trainable parameters by over 99\% compared to full fine-tuning. This efficiency makes GPT4AP attractive for resource-constrained settings, including edge deployment and real-time applications.

Overall, GPT4AP demonstrates the potential of foundation-model-based approaches for environmental time-series forecasting and highlights a promising direction for integrating large pre-trained models into atmospheric science. Future work will focus on incorporating explicit temporal inductive biases into the LLM framework, extending evaluation to broader geographic regions and pollutants, and exploring uncertainty-aware and multi-pollutant forecasting to further enhance the reliability and applicability of the proposed approach.

%------------------------------------------------------------------------------------------

% % use section* for acknowledgement
% \section*{Acknowledgment}
% This research was funded by the Science Foundation Ireland Centre for Research Training in Digitally-Enhanced Reality (d-real) under Grant No. 18/CRT/6224. This research was conducted with the financial support of Science Foundation Ireland under Grant Agreement No.\ 13/RC/2106\_P2 at the ADAPT SFI Research Centre at University College Dublin. ADAPT, the SFI Research Centre for AI-Driven Digital Content Technology, is funded by Science Foundation Ireland through the SFI Research Centres Programme.

% \ifCLASSOPTIONcaptionsoff
%   \newpage
% \fi

\balance

\bibliographystyle{IEEEtran.bst}
\bibliography{./ref/longforms,./ref/references}

\end{document}

\title{Meteorology-Driven GPT4AP: A Multi-Task Forecasting LLM for Atmospheric Air Pollution in Data-Scarce Settings}
%\large Appendix}

%\author{Anonymous Author(s)
\maketitle

% If you want appendix sections labeled as "Appendix A, B, ..."
\appendix

\section{Detailed Ablation Study}

This appendix reports the detailed impact of the rsLoRA rank parameter $R$ on forecasting performance under few-shot, long-term, and zero-shot settings. Tables~\ref{tab:few_shot_rank}-\ref{tab:zero_shot_rank_abl} present station-level and horizon-level results for $R \in \{4, 8, 16, 32, 64\}$. The results consistently show that increasing the rank improves performance up to $R=32$, after which further gains become marginal or inconsistent across tasks and stations. Considering both accuracy and computational efficiency, we adopt $R=32$ as the default configuration for all main experiments.

\subsection{Few-shot Forecasting}

Table~\ref{tab:few_shot_rank} reports few-shot forecasting results using only 10\% of the training data. Increasing the rank from $R=4$ to $R=32$ consistently reduces MSE and MAE across all six stations and all horizons. The overall average MSE decreases from 0.722 at $R=4$ to 0.686 at $R=32$, corresponding to a 5.0\% relative improvement. Increasing the rank further to $R=64$ does not yield additional benefits, with the overall average remaining at 0.686 and the MAE slightly increasing from 0.442 to 0.443. This indicates clear saturation beyond $R=32$ in the few-shot regime.

\begin{table}[!ht]
\centering
\caption{Performance comparison of different rank (R) settings on few-shot forecasting. Values reported as MSE/MAE.}
\label{tab:few_shot_rank}
\scriptsize
\setlength{\tabcolsep}{2pt}
\resizebox{\linewidth}{!}{
\begin{tabular}{l|c|c|c|c|c|c}
\toprule
Dataset & Horizon & R4 & R8 & R16 & R32 & R64 \\
\midrule

\multirow{4}{*}{AZ}
 & 24  &   0.575/0.406 & 0.55/0.392 & 0.550/0.389 & 0.546/0.338 & 0.559/0.387\\
 & 36 &   0.656/0.444 & 0.650/0.438 & 0.632/0.432 & 0.633/0.431 & 0.633/0.430\\
 & 48 &   0.717/0.470 & 0.705/0.463 & 0.694/0.459 & 0.686/0.457 & 0.685/0.456\\
 & 60 &   0.763/0.493 & 0.740/0.482 & 0.741/0.479 & 0.731/0.477 & 0.726/0.475\\
 & Avg & 0.678/0.453 & 0.661/0.444 & 0.654/0.440 & \textbf{0.649}/\textbf{0.426} & \underline{0.651}/\underline{0.437}\\
\midrule

\multirow{4}{*}{DS}
 & 24  &   0.690/0.419 & 0.661/0.405 & 0.663/0.401 & 0.662/0.400 & 0.677/0.398\\
 & 36 &   0.785/0.459 & 0.759/0.450 & 0.762/0.448 & 0.755/0.447 & 0.746/0.444\\
 & 48 &   0.831/0.485 & 0.849/0.481 & 0.819/0.475 & 0.817/0.474 & 0.819/0.474\\
 & 60 &   0.893/0.511 & 0.895/0.503 & 0.868/0.496 & 0.874/0.496 & 0.863/0.494\\
 & Avg & 0.800/0.469 & 0.791/0.460 & 0.778/0.455	& \underline{0.777}/\underline{0.454} & \textbf{0.776}/\textbf{0.453}\\
\midrule

\multirow{4}{*}{SY}
 & 24  &  0.592/0.447 & 0.553/0.423 & 0.540/0.416 & 0.535/0.414 & 0.531/0.411\\
 & 36 &   0.671/0.480 & 0.635/0.463 & 0.622/0.456 & 0.622/0.457 & 0.617/0.454\\
 & 48 &   0.717/0.503 & 0.688/0.488 & 0.681/0.484 & 0.670/0.480 & 0.666/0.476\\
 & 60 &   0.758/0.520 & 0.725/0.504 & 0.714/0.499 & 0.709/0.495 & 0.707/0.495\\
 & Avg & 0.685/0.488 & 0.650/0.470 & 0.639/0.464	& \underline{0.634}/\underline{0.462} & \textbf{0.630}/\textbf{0.459}\\
\midrule

\multirow{4}{*}{TT}
 & 24  &   0.683/0.424 & 0.664/0.410 & 0.685/0.407 & 0.659/0.403 & 0.655/0.401\\
 & 36 &   0.826/0.464 & 0.772/0.454 & 0.741/0.445 & 0.736/0.443 & 0.734/0.441\\
 & 48 &   0.832/0.485 & 0.814/0.474 & 0.800/0.469 & 0.794/0.468 & 0.792/0.466\\
 & 60 &   0.863/0.500 & 0.865/0.494 & 0.846/0.488 & 0.844/0.488 & 0.838/0.485\\
 & Avg & 0.801/0.468 & 0.779/0.458 & 0.768/0.452	& \underline{0.758}/\underline{0.451} & \textbf{0.755}/\textbf{0.448}\\
\midrule

\multirow{4}{*}{HW}
 & 24 &  0.519/0.399 & 0.496/0.383 & 0.496/0.383 & 0.491/0.380 & 0.483/0.374\\
 & 36 &   0.608/0.438 & 0.579/0.426 & 0.579/0.426 & 0.575/0.422 & 0.571/0.419\\
 & 48 &   0.662/0.465 & 0.637/0.452 & 0.637/0.452 & 0.626/0.447 & 0.625/0.445\\
 & 60 &   0.697/0.482 & 0.678/0.472 & 0.678/0.472 & 0.668/0.466 & 0.664/0.464\\
 & Avg & 0.622/0.446 & 0.598/0.433 & 0.598/0.433	& \underline{0.590}/\underline{0.429} & \textbf{0.586}/\textbf{0.426}\\
 \midrule
 
\multirow{4}{*}{WX}
 & 24  &  0.650/0.404 & 0.625/0.392 & 0.613/0.387 & 0.609/0.385 & 0.624/0.384\\
 & 36 &   0.709/0.437 & 0.697/0.431 & 0.700/0.428 & 0.690/0.426 & 0.706/0.427\\
 & 48 &   0.802/0.471 & 0.774/0.458 & 0.748/0.453 & 0.747/0.452 & 0.750/0.452\\
 & 60 &   0.835/0.489 & 0.803/0.477 & 0.797/0.473 & 0.796/0.472 & 0.795/0.472\\
 & Avg & 0.749/0.450 & 0.725/0.440 & 0.715/0.435	& \textbf{0.711}/\textbf{0.434} & \underline{0.719}/\underline{0.434}\\
\midrule
 \multicolumn{2}{c|}{All Avg} & 0.722/0.462 & 0.701/0.451 & 0.692/0.447 & \textbf{0.686}/\textbf{0.442} & \underline{0.686}/\underline{0.443} \\
    \bottomrule
\end{tabular}
}
\end{table}

\subsection{Long-term Forecasting}

Table~\ref{tab:long_term_rank} presents long-term forecasting results using the full training dataset. Performance improves gradually as the rank increases, with the overall average MSE decreasing from 0.680 at $R=4$ to 0.665 at $R=32$ and 0.663 at $R=64$. However, the improvement from $R=32$ to $R=64$ is only 0.3\%, confirming diminishing returns at higher ranks. Similar trends are observed across all stations and horizons, indicating that moderate ranks are sufficient for capturing long-term temporal dependencies.

\begin{table}[!ht]
\centering
\caption{Performance comparison of different rank (R) settings on long-term forecasting. Values reported as MSE/MAE.}
\label{tab:long_term_rank}
\scriptsize
\setlength{\tabcolsep}{2pt}
\resizebox{\linewidth}{!}{
\begin{tabular}{l|c|c|c|c|c|c}
\toprule
Dataset & Horizon & R4 & R8 & R16 & R32 & R64 \\
\midrule

\multirow{4}{*}{AZ}
 & 24  &  0.545/0.387 & 0.536/0.381 & 0.530/0.376 & 0.530/0.376 & 0.524/0.373\\
 & 36 &   0.630/0.430 & 0.624/0.427 & 0.615/0.421 & 0.613/0.419 & 0.611/0.416\\
 & 48 &   0.687/0.457 & 0.677/0.452 & 0.671/0.448 & 0.671/0.447 & 0.666/0.445\\
 & 60 &   0.728/0.476 & 0.719/0.472 & 0.717/0.470 & 0.712/0.467 & 0.709/0.465\\
 & Avg & 0.648/0.438 & 0.639/0.433 & 0.633/0.429	& \underline{0.632}/\underline{0.427} & \textbf{0.628}/\textbf{0.425}\\
\midrule

\multirow{4}{*}{DS}
 & 24  &  0.650/0.398 & 0.641/0.393 & 0.636/0.388 & 0.636/0.389 & 0.644/0.391\\
 & 36 &   0.744/0.444 & 0.736/0.439 & 0.731/0.434 & 0.726/0.432 & 0.728/0.433\\
 & 48 &   0.806/0.473 & 0.798/0.468 & 0.792/0.463 & 0.790/0.462 & 0.788/0.460\\
 & 60 &   0.855/0.494 & 0.844/0.488 & 0.843/0.487 & 0.839/0.484 & 0.836/0.481\\
 & Avg & 0.764/0.452 & 0.755/0.447 & 0.751/0.443	& \textbf{0.748}/\underline{0.442} & \underline{0.749}/\textbf{0.441}\\
\midrule

\multirow{4}{*}{SY}
 & 24  &  0.521/0.403 & 0.513/0.398 & 0.508/0.392 & 0.508/0.392 & 0.505/0.390\\
 & 36 &   0.608/0.447 & 0.599/0.441 & 0.594/0.438 & 0.592/0.435 & 0.592/0.435\\
 & 48 &   0.661/0.473 & 0.653/0.468 & 0.646/0.463 & 0.645/0.461 & 0.643/0.460\\
 & 60 &   0.701/0.491 & 0.694/0.487 & 0.688/0.483 & 0.684/0.480 & 0.683/0.479\\
 &  Avg & 0.623/0.454 & 0.615/0.449 & 0.609/0.444 & \underline{0.607}/\underline{0.442} & \textbf{0.606}/\textbf{0.441}\\
\midrule

\multirow{4}{*}{TT}
 & 24  &   0.643/0.397 & 0.636/0.393 & 0.631/0.388 & 0.632/0.388 & 0.627/0.385\\
 & 36 &   0.735/0.441 & 0.727/0.436 & 0.721/0.430 & 0.715/0.427 & 0.722/0.430\\
 & 48 &   0.793/0.465 & 0.785/0.461 & 0.780/0.457 & 0.778/0.456 & 0.773/0.453\\
 & 60 &   0.837/0.485 & 0.827/0.480 & 0.824/0.477 & 0.821/0.475 & 0.819/0.472\\
 & Avg & 0.752/0.447 & 0.744/0.443 & 0.739/0.438	& \underline{0.737}/\underline{0.437} & \textbf{0.735}/\textbf{0.435}\\
\midrule

\multirow{4}{*}{HW}
 & 24  &   0.490/0.379 & 0.480/0.372 & 0.480/0.372 & 0.473/0.362 & 0.470/0.363\\
 & 36 &   0.572/0.420 & 0.565/0.416 & 0.565/0.416 & 0.557/0.401 & 0.555/0.406\\
 & 48 &   0.625/0.445 & 0.616/0.441 & 0.616/0.441 & 0.612/0.422 & 0.610/0.434\\
 & 60 &   0.670/0.466 & 0.658/0.460 & 0.658/0.460 & 0.654/0.432 & 0.650/0.453\\
 & Avg & 0.589/0.428 & 0.580/0.422 & 0.580/0.422	& \underline{0.574}/\textbf{0.404} & \textbf{0.571}/\underline{0.414}\\
 \midrule
 
\multirow{4}{*}{WX}
 & 24  &  0.603/0.382 & 0.595/0.378 & 0.591/0.373 & 0.588/0.372 & 0.592/0.373\\
 & 36 &  0.688/0.426  & 0.681/0.422 & 0.675/0.416 & 0.671/0.413 & 0.672/0.412\\
 & 48 &  0.744/0.453  & 0.737/0.448 & 0.733/0.445 & 0.731/0.442 & 0.726/0.440\\
 & 60 &  0.790/0.473  & 0.779/0.467 & 0.775/0.463 & 0.779/0.464 & 0.772/0.460\\
 & Avg & 0.706/0.434 & 0.698/0.429 & 0.694/0.424	& \underline{0.692}/\underline{0.423} & \textbf{0.691}/\textbf{0.421}\\
 \midrule
 \multicolumn{2}{c|}{All Avg} & 0.680/0.442 & 0.672/0.437 & 0.668/0.433 & \underline{0.665}/\textbf{0.429} & \textbf{0.663}/\underline{0.430}\\
\bottomrule
\end{tabular}
}
\end{table}

\subsection{Zero-shot Forecasting}

\begin{table}[!ht]
\centering
\caption{Performance comparison of different rank (R) settings on zero-shot forecasting. Values are reported as MSE/MAE.}
\scriptsize
\setlength{\tabcolsep}{2pt}
\begin{tabular}{l|c|c|c|c|c|c}
\toprule
Dataset & Horizon & R4 & R8 & R16 & R32 & R64 \\
\midrule

\multirow{4}{*}{\makecell{AZ \\ $\rightarrow$ \\ DS}}
 & 24  & 0.465/0.364 & 0.457/0.351 & 0.455/0.348 & 0.455/0.351 & 0.453/0.350 \\
 & 36  & 0.509/0.389 & 0.506/0.385 & 0.501/0.382 & 0.502/0.380 & 0.504/0.381 \\
 & 48  & 0.555/0.414 & 0.542/0.405 & 0.543/0.403 & 0.541/0.403 & 0.547/0.406 \\
 & 60  & 0.592/0.438 & 0.582/0.426 & 0.585/0.427 & 0.583/0.424 & 0.583/0.424 \\
 & Avg & 0.530/0.401 & 0.522/0.392 & \underline{0.521}/\underline{0.390} & \textbf{0.520}/\textbf{0.390} & 0.522/0.390\\
\midrule

\multirow{4}{*}{\makecell{DS \\ $\rightarrow$ \\ AZ}}
 & 24  & 0.457/0.386 & 0.439/0.373 & 0.447/0.376 & 0.420/0.366 & 0.452/0.376 \\
 & 36  & 0.504/0.411 & 0.502/0.409 & 0.490/0.402 & 0.495/0.401 & 0.479/0.401 \\
 & 48  & 0.560/0.442 & 0.532/0.424 & 0.527/0.424 & 0.525/0.422 & 0.539/0.426 \\
 & 60  & 0.587/0.456 & 0.605/0.453 & 0.591/0.448 & 0.581/0.445 & 0.575/0.443 \\
 & Avg & 0.527/0.424 & 0.520/0.415 & 0.514/0.413	& \textbf{0.505}/\textbf{0.409} & \underline{0.511}/\underline{0.412}\\
\midrule

\multirow{4}{*}{\makecell{AZ \\ $\rightarrow$ \\ TT}}
 & 24  & 0.499/0.378 & 0.489/0.366 & 0.485/0.362 & 0.484/0.364 & 0.484/0.365 \\
 & 36  & 0.540/0.402 & 0.533/0.398 & 0.528/0.394 & 0.527/0.392 & 0.531/0.394 \\
 & 48  & 0.584/0.426 & 0.564/0.415 & 0.564/0.413 & 0.562/0.414 & 0.566/0.416 \\
 & 60  & 0.617/0.448 & 0.598/0.435 & 0.602/0.436 & 0.600/0.433 & 0.601/0.433 \\
 & Avg & 0.560/0.414 & 0.546/0.404 & \underline{0.545}/\underline{0.401}	& \textbf{0.543}/\textbf{0.401} & 0.546/0.402\\
\midrule

\multirow{4}{*}{\makecell{TT \\ $\rightarrow$ \\AZ}}
 & 24  & 0.516/0.423 & 0.485/0.394 & 0.464/0.385 & 0.450/0.376 & 0.446/0.375 \\
 & 36  & 0.543/0.430 & 0.505/0.411 & 0.512/0.414 & 0.497/0.407 & 0.508/0.411 \\
 & 48  & 0.575/0.447 & 0.563/0.436 & 0.549/0.431 & 0.549/0.430 & 0.569/0.440 \\
 & 60  & 0.629/0.471 & 0.605/0.458 & 0.604/0.456 & 0.573/0.444 & 0.575/0.445 \\
 & Avg & 0.566/0.443 & 0.540/0.425 & 0.532/0.422	& \textbf{0.517}/\textbf{0.414} & \underline{0.525}/\underline{0.418}\\
\midrule

\multirow{4}{*}{\makecell{DS \\ $\rightarrow$ \\TT}}
 & 24  & 0.513/0.385 & 0.501/0.373 & 0.503/0.376 & 0.489/0.368 & 0.513/0.376 \\
 & 36  & 0.549/0.406 & 0.548/0.406 & 0.538/0.399 & 0.546/0.398 & 0.532/0.400 \\
 & 48  & 0.599/0.436 & 0.577/0.419 & 0.569/0.419 & 0.572/0.417 & 0.576/0.420 \\
 & 60  & 0.615/0.448 & 0.630/0.445 & 0.624/0.441 & 0.617/0.438 & 0.610/0.436 \\
 & Avg & 0.569/0.419 & 0.564/0.411 & 0.559/0.409	& \textbf{0.556}/\textbf{0.405} & \underline{0.558}/\underline{0.408}\\
\midrule

\multirow{4}{*}{\makecell{TT \\ $\rightarrow$ \\DS}}
 & 24  & 0.514/0.403 & 0.501/0.377 & 0.494/0.369 & 0.477/0.360 & 0.473/0.360 \\
 & 36  & 0.539/0.409 & 0.516/0.393 & 0.523/0.395 & 0.520/0.392 & 0.526/0.394 \\
 & 48  & 0.568/0.427 & 0.572/0.418 & 0.559/0.413 & 0.560/0.412 & 0.574/0.421 \\
 & 60  & 0.618/0.450 & 0.607/0.439 & 0.602/0.436 & 0.581/0.426 & 0.582/0.427 \\
 & Avg & 0.560/0.422 & 0.549/0.407 & 0.545/0.403	& \textbf{0.535}/\textbf{0.398} & \underline{0.539}/\underline{0.401}\\
 \midrule
 \multicolumn{2}{c|}{All Avg} & 0.552/0.420 & 0.540/0.409 & 0.536/0.406 & \textbf{0.529}/\textbf{0.403} & \underline{0.533}/\underline{0.405}\\
\bottomrule
\end{tabular}
\label{tab:zero_shot_rank_abl}
\end{table}
Table~\ref{tab:zero_shot_rank_abl} reports zero-shot cross-station transfer results, where models trained on one station are evaluated on another without fine-tuning. The overall average MSE decreases from 0.552 at $R=4$ to 0.529 at $R=32$, corresponding to a 4.2\% relative improvement. Increasing the rank to $R=64$ slightly degrades performance to 0.533, suggesting mild overfitting or reduced generalization at higher ranks. These results indicate that $R=32$ provides the best balance between expressiveness and generalization in transfer settings.

Overall, across all forecasting regimes, $R=32$ achieves near-optimal accuracy while maintaining strong parameter efficiency. This configuration consistently offers the best trade-off between forecasting performance, generalization capability, and computational cost.

% If appendix contains citations, include the same bibliography:
%\bibliographystyle{IEEEtran.bst}
%\bibliography{./ref/longforms,./ref/references}